\documentclass{article}

\usepackage{PRIMEarxiv}

\usepackage[utf8]{inputenc} % allow utf-8 input
\usepackage[T1]{fontenc}    % use 8-bit T1 fonts
\usepackage{hyperref}       % hyperlinks
\usepackage{amsmath}
\usepackage{bbm}
\usepackage{url}            % simple URL typesetting
\usepackage{booktabs}       % professional-quality tables
\usepackage{amsfonts}       % blackboard math symbols
\usepackage{nicefrac}       % compact symbols for 1/2, etc.
\usepackage{microtype}      % microtypography
\usepackage{lipsum}
\usepackage{fancyhdr}       % header
\usepackage{graphicx}       % graphics
\graphicspath{{media/}}     % organize your images and other figures under 
\usepackage{import}

\title{IMAD: IMage-Augmented multi-modal Dialogue
}

\author{
  Viktor Moskvoretskii \\
  DeepPavlov.ai \\
  \texttt{vvmoskvoretskiy@yandex.ru} \\
   \And
  Anton Frolov \\
  DeepPavlov.ai \\
  \texttt{antonylovanto@gmail.com} \\
  \And
  Denis Kuznetsov \\
  DeepPavlov.ai \\
  \texttt{kuznetsov.den.p@gmail.com} \\
}

\begin{document}

\maketitle

\begin{abstract}
  Currently, dialogue systems have achieved high performance in processing text-based communication. However, they have not yet effectively incorporated visual information, which poses a significant challenge. Furthermore, existing models that incorporate images in dialogue generation focus on discussing the image itself. Our proposed approach presents a novel perspective on multi-modal dialogue systems, which interprets the image in the context of the dialogue. By doing so, we aim to expand the capabilities of current dialogue systems and transition them from single modality (text) to multi-modality. However, there is a lack of validated English datasets that contain both images and dialogue contexts for this task. Thus, we propose a two-stage approach to automatically construct a multi-modal dialogue dataset. In the first stage, we utilize text-to-image similarity and sentence similarity to identify which utterances could be replaced with an image. In the second stage, we replace those utterances by selecting a subset of relevant images and filtering them with a visual question answering model. We used this approach, along with additional labeling, to create the IMage Augmented multi-modal Dialogue dataset (IMAD), which can serve as a validated dataset for this task. Furthermore, we propose a baseline model trained on this dataset, which outperforms model trained on the same data without images and BlenderBot.
\end{abstract}

% keywords can be removed

\section{Introduction}
Dialogue systems, also known as conversational agents or chatbots, have become increasingly important in recent years due to their potential to revolutionize human-computer interaction \cite{dialogue-systems-review}.% By delegating mundane and repetitive tasks, dialogue systems can assist human agents in complex issues while reducing costs \cite{yan2017building}. 
Furthermore, one can see a high activity in this field in the recent year, as the usage of ChatGPT \cite{chat-gpt} serves for a lot of different goals \cite{chat-gpt-usage,human-computer-interaction,lund2023chatting}. Just like with the ChatGPT recently Google has announced Bard, that is based on LaMDA \cite{lamda}, which serves for the same tasks.
Additionally, dialogue systems provide a challenging problem in AI research, as they require a deep understanding of natural language and the ability to generate human-like responses \cite{Chen_2017}. A good confirmation of this thesis is the abundance of different natural language products, that are widely used, such as DeepPavlov \cite{burtsev2018deeppavlov}.

\smallskip

Contemporary dialogue models such as DialoGPT, BLOOM, and DialogBERT are predominantly text-based \cite{zhang2019dialogpt,Scao-2022,Gu-2020}. This is reasonable in scenarios where individuals converse solely through textual communication. However, in real-life situations, dialogues frequently incorporate images, such as when individuals respond to questions with photographs, provide offers or  express emotions \cite{mm_chat}. As a result, there is a need for dialogue models that can accommodate multi-modal inputs. 

\smallskip

Just as in the dialogues systems high activity is being spotted in the field of text2image generation \cite{imagen,stable-diffusion,Dalle}. These tools are also helpful in the art sphere \cite{AI-art}, healthcare \cite{chen2022generative}, physics \cite{manyar2023physics} and more. However, these models are limited to producing only images and taking text information as input.

\medskip

These problems are solved with multi-modal deep learning models, that are increasingly important today due to the exponential growth of multimedia data in various domains \cite{Tadas_multimodal,mulit-modal-review,multi-modal-explainability,multi-modal-fusion}.
Such models have numerous applications in various domains. %In e-commerce, these models can improve product recommendation systems by considering both the textual description and visual features of products \cite{Yu_2022}. 
There is a strong potential of incorporating more modalities to dialogue assistants, such as better emotion recognition \cite{emotion-multimodal}, visual question answering \cite{vqa-multimodal,blip} and operating with the wide range of tasks \cite{GATO,flamingo,kosmos-1}.

\smallskip

Therefore, one could focus on the task of generating an image description in a context of dialogue \cite{Lee-2021,mm_chat}. This task is more general compared to the response generation \cite{chat-gpt} or describing a picture \cite{blip2}, because it allows for better response generation with intention knowledge and knowledge about the image sense in a certain dialogue context.

\smallskip

\textit{Proposal.} That brings us to the task of interpreting an image in the context of dialogue, that is exactly the solution of aforementioned problem. We present IMage Augmented multi-modal Dialogue dataset (IMAD), that contains 4864 dialogues, where last utterance was replaced with image. To collect it we utilize multiple sources of dialogue dataset, present novel approach for image-text dialogue construction and labeled part of it with 3 assessors. We also present baseline models, based on the BLIP \cite{blip}, that outperform text-only BLIP and BlenderBot 400M  \cite{blenderbot} on this task. Train data included 4154 samples collected with automated approach and 582 samples labeled with assessors as "Partial Match". Test data included 128 samples, labeled by assessors and authors as "Perfect Match". 

\smallskip

Out dataset and code are published at \href{https://github.com/VityaVitalich/IMAD}{IMAD Repository}

\section{Related Works}
\textbf{Multi-modal Tasks}. Multi-modal models involve multiple modalities, such as images \cite{blip,Dalle,sim-vlm,VinVl}, video \cite{frozen,video-qa,VideoCoCa}, or audio \cite{flow-tron,natural-speech}. In the field of text and image modalities there are variety of popular tasks. Visual question answering \cite{blip2,blip,ALBEF,ALIGN}, Image captioning \cite{kumar2022imagecaptioning,OFA,mPLUG}. As well, we distinguish as separate task image-text matching, that key is to match corresponding images and texts \cite{Radford-2021,blip,bridge-towers}

\textbf{Mutli-modal Embeddings}. Modern models \cite{GATO,kosmos-1,flamingo} focus on connecting multiple modalities in one model. In contrast, we would like to focus on text with images data. One of the strongest multi-modal model was CLIP \cite{Radford-2021}, which uses image embeddings to align with text embeddings for image-text matching loss \cite{contrastive-loss}, so images will correspond to relevant phrases. The same idea was used in BLIP \cite{blip} with matching image and text embeddings \cite{ALBEF}. This is a key idea to filter pairs of text and image.

\textbf{Multi-modal Data}. Current multi-modal data contains images with captions \cite{laion,SBU-captions,visual-genome,conceptual,ms-coco,ALIGN}. BLIP, BLIP2 and Flamingo were trained on these. However, these datasets do not contain dialogue contexts.

Previous research has also encountered a similar challenge, as evidenced by a study on the topic of image-grounded dialogues \cite{mm_chat}. The authors of this study utilized dialogues from Chinese social media and crowd-sourcing to develop their model. The authors reported an increase in BLEU \cite{bleu} scores compared to generated responses that were not conditioned on image data.

Another study \cite{Lee-2021} proposes constructing a dataset utilizing image-text matching via the Visual Semantic Reasoning Network (VSRN) \cite{li2019vsrn} with images sourced from the MS COCO \cite{ms-coco} and Flicker 30k \cite{flicker30k} datasets, just as it was done before \cite{dialogcc}.

\section{Dialogue Datasets}
To produce clean data, it is important to have dialogues created with humans. They could be collected either from crowd-sourcing or written by humans, such as in English books. Therefore, we have chosen the dataset sources listed below to make our data valid and diverse in terms of dialogue content: DailyDialog, Persona Chat, MuTual, DREAM, Common-sense dialogues, and Empathetic Dialogues \cite{Zhang-2018,Li-Su-2017,cui-etal-2020-mutual,Zhou-2021,Rashkin-2018,Sun-2019}. Detailed reasons are provided in Appendix B.

\smallskip

With these sources, we have collected 451,611 pairs of utterances with context and images. This dataset contains all the features and model predictions described below.

\smallskip

The filtered version of the dataset is named IMage-Augmented Multi-modal Dialogue Dataset (IMAD) and is used later for modeling. It is constructed with a combination of the Text-Image Replacing approach and Human Annotated part. Basic statistics are shown in Table \ref{TableBasicStatistics}.

\section{Text-Image Replacing}

Previously approach with filter from VSRN did not show well results \cite{Lee-2021}. 
Therefore to obtain more precise and appropriate dataset, we utilized a two-step process to determine the feasibility of replacing an utterance with an image. The first step involved predicting whether the utterance could be substituted with an image. The second step focused on matching a better picture utilizing VQA from BLIP.

\subsection{Find Replaceable Utterances}
% В этом разделе надо:
% - коротко рассказать, что мы в начале нагенерили призаков описывающих замену, пназвали их скорами. Разметили данные. Построили бустинг
% - описать сами признаки замены
% - описать разметку для бустинга
% - сам бустинг+выводы с фичеинпотансом

The first step is to find utterances that could be replaced with an image. To accomplish this task, we have labeled a little subset, created features (we will name them scores) and built Random Forest model \cite{scikit-learn}.

\subsubsection{Human Annotation.}

For the initial step, we labeled 1000 random samples from the DailyDialogue \cite{Li-Su-2017}, state them as $U = \{u_1 \dots u_{1000}\}$, that are utterances with contexts. To label them, we were using a heuristic formulated as "\textit{This phrase potentially can be described with a picture}".

\subsubsection{Replacing Features.}

The matching process involves pairing each utterance with an image from the Unsplash. This is achieved by maximizing the cosine similarity between the embeddings of the utterance and the image extracted from CLIP \cite{Radford-2021}. 

\smallskip

To optimize the matching process, we conducted experiments using various text and image features that were deemed important for predicting if utterance is replaceable. The most promising results were obtained when each utterance was accompanied by the following features: Image Score, Maximum Entity Score, Sentence Similarity, BLEU Score, Threshold. Detailed description is provided in Appendix A.

\smallskip

\subsubsection{ML Labeling.}

For classification we employed the Random Forest algorithm, that demonstrated the best precision\cite{scikit-learn}. This behavior is likely attributed to the high variance in the data, as evidenced by the standard deviation.

Multiple tests were conducted with the Stratified K-fold cross-validation with 3 folds and 40 repeats. Precision was deemed to be the key metric, given the importance of minimizing errors, as even the exclusion of valid utterances is preferable to making errors.

\smallskip

\begin{figure}[h]
\centering
% TODO: uncomment
 \includegraphics[width=8cm]{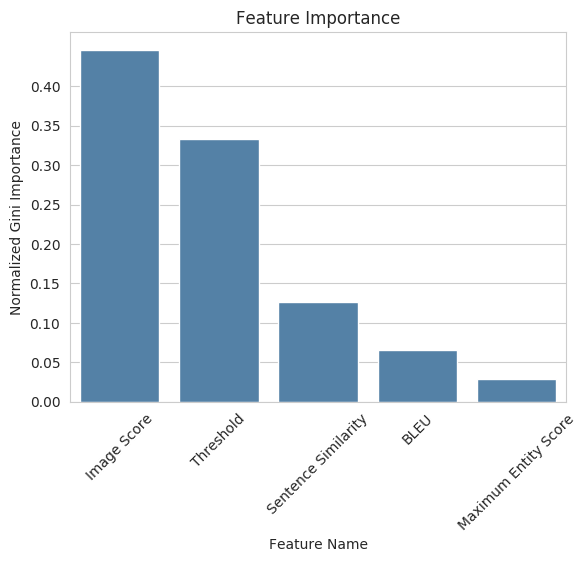}
\caption{Feature Importance in Random Forest Model for predicting if an utterance could be replaced with an image. Image Score is maximum cosine similarity between utterance and images embeddings from CLIP. Threshold is binary feature, indicating if Image Score, Sentence Similarity and Maximum Entity Score is bigger than empirically founded values. Sentence Similarity is cosine distance similarity between utterance embedding and caption embedding, that were generated from corresponding image with VIT-GPT2, where embeddings come from SentenceBert. BLEU is calculated between utterance and generated caption with taking only unigrams into account. Maximum Entity Score is maximum out of Image Scores for each noun in utterance.}
\label{pictorialization::FI}
\end{figure}

The resulting model metrics are shown in Table~\ref{TableMetricsRF} and feature importances are shown in Fig.~\ref{pictorialization::FI}.

\subsection{Text-Image Matching}
% \subsection{VQA Text-Image Matching}

The criterion from the first step, which considers only the text and not the image, serves as a primary filter for the text-image task. However, it is insufficient on its own, that is shown by not so high correlation in previous works \cite{Lee-2021}. The image dataset limitations impose constraints on the ability to form a pair of utterance and image, even if the utterance is replaceable. To overcome this, it is necessary to introduce a step to match utterances with better images.

\subsubsection{ML Labeling.}

In order to improve the quality of the images, we utilized the BLIP VQA \cite{li2022blip}. To select proper images we use confidence of the model output, which is defined as sum of the log probabilities of each token in the utterance. $\operatorname{Confidence}(out, u) = \sum\limits_{i=1}^{length \, token(u)} out_{i, token(u)_i}$.

\smallskip

The process of selecting a better image for an utterance was carried out using the following steps:
\begin{enumerate}
    \item \textbf{Create scoring for all images}. For each utterance we have cosine similarity with every image in dataset $ \Bigl\{ \operatorname{cosine}(\operatorname{emb}_{CLIP}(u), \operatorname{emb}_{CLIP}(i)) \quad \forall i \Bigr\}   =: ISS_u$.
    
    \item \textbf{Create set of N images}. From all that set we take set of N images, which are the top-N for cosine similarity $\Bigl\{i \, | \, \operatorname{cosine}(\operatorname{emb}_{CLIP}(u), \operatorname{emb}_{CLIP}(i)) \geq ISS_{u_{(N)}} \Bigr\} =: TopImg_{u, N}$
    %, where $ISS_{u_{(N)}}$ defines N order value in $ISS_u$ sorted in descending order.

    \item \textbf{Query the VQA}. The model was queried with the text input "Which phrase can describe this image?" for each image in the aforementioned set.
    
    % $$\forall u \quad \forall i \in TopImg_{u, N} \quad \exists VO_{quest, i} = \textit{VQA Output for image} = VQA(quest, i)$$,

    % where $VQA(quest, i)$ is output of VQA model for question $quest$ and image $i$

    \item \textbf{Calculate confidences}. For each image in aforementioned set we calculate confidence $\Bigl\{ \operatorname{Confidence(out, u)} \, \big| \, out = \operatorname{VQA}(quest, i) \quad \forall i \in TopImg_{u,N}\Bigr\} =: ConfSet_{u,N} $.

    \item \textbf{Select the most confident}. Then we select image with the highest confidence score $\operatorname{argmax} \{ConfSet_{u,N}\} =: img_{N}$.
\end{enumerate}

\smallskip

We conducted a test of our methodology by labeling image-text matching in the context of dialogue pairs that were previously identified as replaceable. The labeling process involved 3 classes: "\textit{Image matches}", "\textit{Image does not match}" and "\textit{Unknown}" in cases where determination was difficult. Our results, as presented in Table~\ref{TableVQA}, confirm that our initial assumptions were correct. Specifically, our findings indicate that, at best, only half of the pairs with replaceable utterances were found to have matching images.

\subsubsection{Human Annotation.}

%To refine the dataset, a RF model was employed to identify utterances that could be effectively replaced with corresponding images. The selected images were further matched with the textual context using a Visual Question Answering (VQA) model from BLIP \cite{blip} to improve the accuracy of the image selections. However, due to low recall, only 
With the above method we obtained a subset of 4154 samples. In order to enlarge the dataset, samples with lower RF scores were selected and labeled by three expert assessors, resulting in the addition of 4644 more samples to the dataset. The labeled dataset was organized into 4 distinct categories: "\textit{Perfect Match}", "\textit{Partial Match}", "\textit{Undefined}" and "\textit{No Match}". Detailed description and labeling instructions provided  in Appendix D.

\smallskip

The inter-rater reliability of their annotations were evaluated using Fleiss-kappa statistic. The refined version of the IMage Augmented multi-modal Dialogue dataset (IMAD) is constructed from samples obtained with our initial approach and samples from dataset labeled with assessors, that had label "Perfect Match" or "Partial Match". Basic statistics are shown in Table~\ref{TableBasicStatistics}, number of samples from different sources and Fleiss's Kappa are shown in Table~\ref{TableAccuracies}.

\begin{table}
\caption{Accuracy for each data source for each label in assessor's validation dataset, where ground true labeling were done with authors. Fleiss Kappa across assessors for each data source. Number of samples in resulting dataset from each data source.}
\label{TableAccuracies} 
\begin{center}
\tabcolsep=0.11cm
\begin{tabular}{l|cccc|c|c}
\hline
Dataset &    1 class &     2 class &     3 class &    4 class & Fleiss Kappa & \# of samples\\
\hline
PersonaChat         &  0.29 &  0.43 &  0.82 &  1.0 & 0.83 & 2483  \\
DailyDialog        & - &  0.50 & - & - & 0.80 & 899 \\
EmpatheticDialogues   & - & - & - & - & 0.76 & 754 \\
Commonsense-Dialogues & - & - &  1.0 & - & 0.83 & 220\\
MuTual        & - & - & - & - & 0.88 & 333\\
DREAM        & - & - & - & - & 0.81 & 175 \\
\hline
Mean Across All Sources & 0.29 & 0.46 & 0.9 & 1.0 & 0.82 & 810.67 \\
\hline
\end{tabular}
\end{center}

\end{table}

\begin{table}[htb]
    \parbox{.5\linewidth}{
        \caption{Number of image-text matches for different N in VQA image-text matching approach}
        \label{TableVQA}
        \centering
        \tabcolsep=0.11cm
            \scalebox{0.8}
        {
        \begin{tabular}{lccc}
        \hline N & Image Matches & No Match &  Unknown \\ \hline
        1  & 37 & 51 & 8\\
        5 & 46 & 46 & 4 \\
        10 & 48 & 44 & 4 \\
        15 & 47 & 45 & 4 \\
        50 & 42 & 51 & 3 \\
        \hline
        \end{tabular}
        } 
       
    }
    \hfill
    \parbox{.5\linewidth}{
        \caption{Basic Statistics of Dataset}
        \label{TableBasicStatistics} 
        \centering
        \tabcolsep=0.11cm
        \scalebox{0.8}
        {
        \begin{tabular}{lc}
        \hline
        Total Dialogues         &         4864 \\
        Average Speaker Turns Per Context         &          5.1 \\
        Average Number of Tokens Per Context   &          56.4 \\
        Average Number of Tokens Per Replaced Utterance       &          14.5 \\
        Size of Context Vocabulary & 12375 \\
        Size of Replaced Utterances Vocabulary &   7962 \\
        \hline
        \end{tabular}
        }    
        }
\end{table}

\section{Multi-modal Dialogue Language Model}
To validate our approach we train a model on the proposed dataset using both image and text data and compare it to text-only models. We choose the task of reconstructing the substituted utterance as visual signal is clearly beneficial in such setting.

\subsection{Model architecture}

We choose a pre-trained BLIP \cite{blip} model for experiments as it is one of the best open source models utilising both visual and text modalities and has a handy interface in LAVIS library \cite{lavis}. The model consists of a visual transformer \cite{vit} for image encoding and a BERT \cite{bert} with a language modeling head for text decoding, both initialised from a BLIP checkpoint. Training details shown in Appendix C.

\medskip

To validate the composed dataset we finetuned two BLIP models from a pre-trained checkpoint. These models consist of ViT-B/16 image encoder and BERT text decoder with 12 layers, 12 attention heads, hidden size of 768, intermediate size of 3072, and GeLU activations \cite{gelu}. Total number of parameters is 224M.

\subsection{Evaluation}

For evaluation we use 128 well annotated samples by both us and the assessors. For each trained model we choose the best checkpoint using validation metrics and use it to compute metrics on the test split. We compare BLIP finetuned on the proposed dataset to zero-shot performance of a distilled BlenderBot 400M \cite{blenderbot} and BLIP finetuned in text-only setting. We choose BLEU \cite{bleu} with n-grams lengths from 1 to 4 as quality metrics and also report perplexity for the models we trained. During evaluation we use beam search sampling with 3 beams. We also divide the test split of the dataset into parts corresponding to the source dialog corpuses and report the metrics for each of them in Table \ref{table:test_metrics_by_source} and for the whole set in Table \ref{table:test_metrics}.

\subsection{Generation examples}

\begin{figure}
    \centering
    % TODO: uncomment
    \includegraphics[width=0.49\linewidth, viewport=0 0 2154 1464]{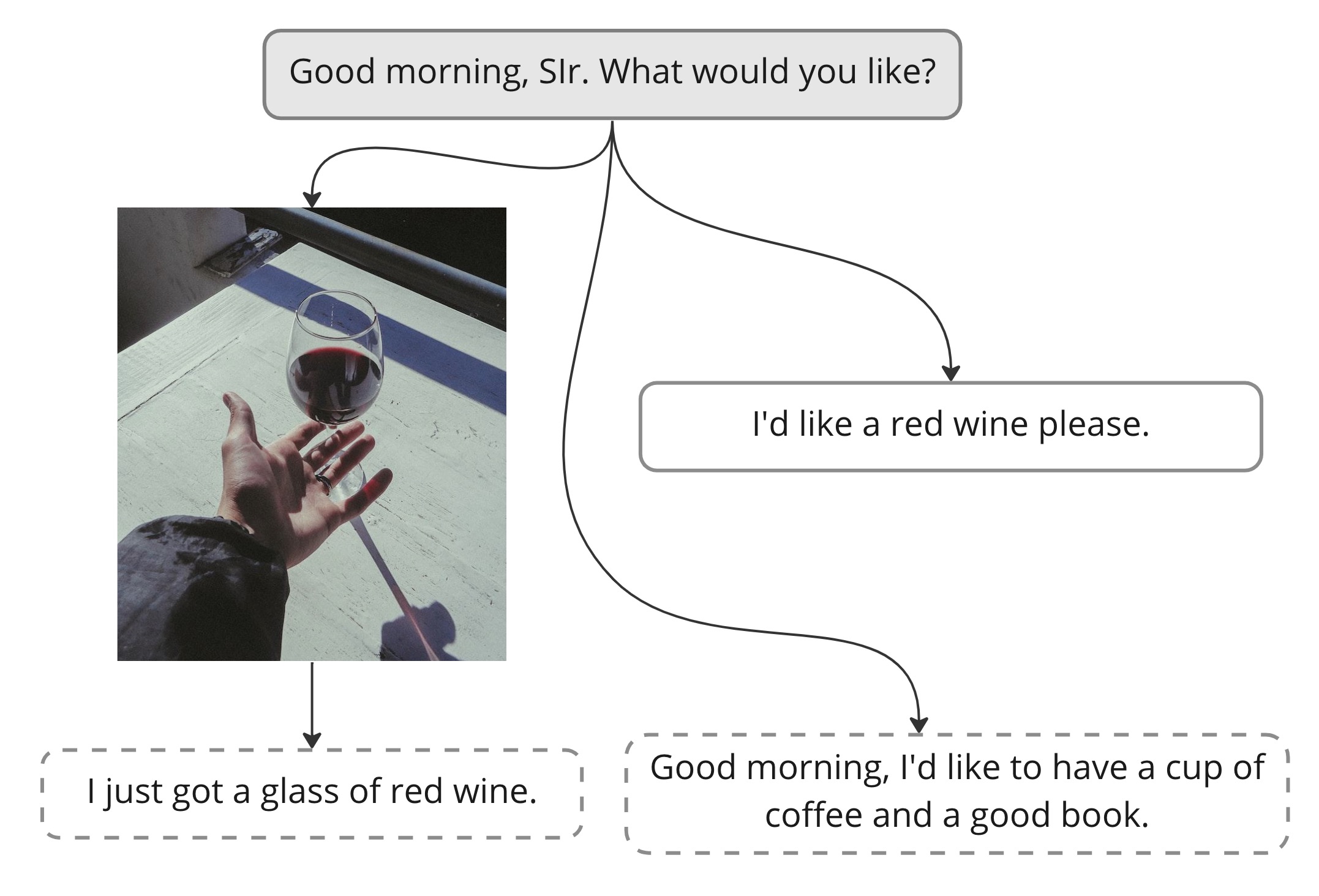}
    \includegraphics[width=0.49\linewidth, viewport=0 0 1728 966]{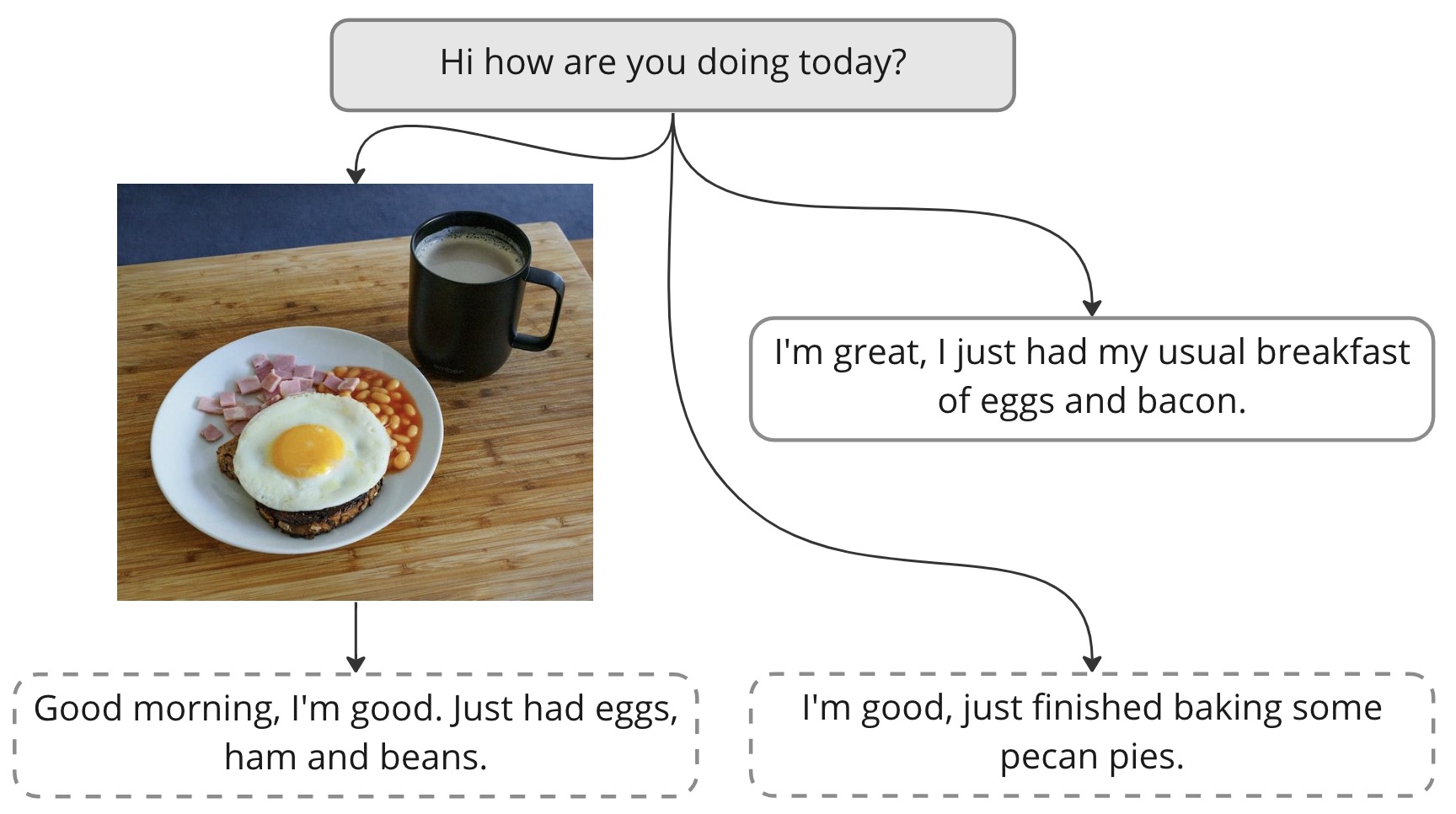}
    \caption{Examples of finetuend models generation: grey blobs represent the context, white blobs represent ground truth utterances, and dashed blobs represent model generation outputs with or without using visual input.}
    \label{fig:examples}
\end{figure}

Figure \ref{fig:examples} show some examples of model generation results. We find that the model finetuned using visual data uses that information in its answers as opposed to the one finetuned using constant visual inputs. As we have only one ground truth label for one sample sometimes model outputs do not exactly align with them, but usually make sense and come close.

\begin{table}
\caption{Test split metrics for both finetuned models and BlenderBot 400M.}
\label{table:test_metrics}
\centering
\tabcolsep=0.11cm
\scalebox{0.9}{
\begin{tabular}{lccccc}
\toprule
{} &  BLEU-1 &  BLEU-2 &  BLEU-3 &  BLEU-4 &  Perplexity \\\midrule
image+text & \textbf{23.73} $\pm$ 1.25 & \textbf{14.37} $\pm$ 1.06 & \textbf{9.19} $\pm$ 0.84 & \textbf{6.33} $\pm$ 0.77 & \textbf{44.19} $\pm$ 1.00 \\
text-only  & 10.63 $\pm$ 0.60 & 5.76 $\pm$ 0.35 & 4.01 $\pm$ 0.28 & 3.23 $\pm$ 0.24 & 90.21 $\pm$ 1.04 \\
BlenderBot & 10.93 & 4.75 & 2.62 & 1.61 & - \\
\bottomrule
\end{tabular}
}
\end{table}

\begin{table}[pt]
\caption{Metrics on test split by source for both finetuned models and BlenderBot 400M. We report metrics for top-3 most frequent sources.}
\label{table:test_metrics_by_source}
\centering
\scalebox{0.9}{
\begin{tabular}{lcccccc}
\toprule
{} & \multicolumn{2}{c}{Persona-Chat} & \multicolumn{2}{c}{DailyDialog} & \multicolumn{2}{c}{EmpatheticDialogues} \\
{} &        BLEU-1 &     BLEU-4  &        BLEU-1 &     BLEU-4 &       BLEU-1 &     BLEU-4 \\\midrule
image+text &\textbf{22.81} & \textbf{5.52} &       \textbf{26.51} & \textbf{6.27} &             \textbf{27.64} & \textbf{9.23} \\
text-only  &       9.73 & 2.68 &       13.81 & 3.79 &             8.98 & 0.00 \\
BlenderBot &       13.29 & 2.83 &       12.17 & 0.00 &             10.24 & 1.62 \\
\bottomrule
\end{tabular}
}
\end{table}

\section{Results and Discussion}

\textbf{Dataset}. We have presented a dataset (IMAD) with a considerable amount of multi-modal dialogues, sourced from validated text-only datasets. The creation of our dataset is automated with a two-step process, involving the filtering of the most relevant utterances and the selection of the most suitable image. Additionally, we developed a methodology for labeling the data, which proved to be well-understood and valid. This is demonstrated by the high Fleiss Kappa score, which measures the consistency between 3 assessors, as shown in Table ~\ref{TableAccuracies}. The basic statistics of our dataset are presented in Table ~\ref{TableBasicStatistics}, and the statistics per dialogue are quite similar to those of DailyDialogue, which is a well-validated dataset.

\smallskip

\textbf{Baseline Model}. In this study, we propose a model for the task of generating an utterance that is replaced with an image, using the IMAD. It is based on the BLIP architecture and achieves a relatively high BLEU score compared to other models that use only text information (as shown in Table~\ref{table:test_metrics}). This result demonstrates the validity of our dataset, which is consistent across different sources of data (as shown in Table~\ref{table:test_metrics_by_source}). We have overcame the issue of noisy and irrelevant pairs of utterances and images by incorporating a filtering stage and creating the IMAD, resulting in a model that potentially could outperform a previous approach due to cleaner data \cite{mm_chat}.

\smallskip

\textbf{Further Work}. These promising findings are expected to significantly contribute to the advancement of research in the field of multi-modal dialogue models. The methodology presented in this paper could aid researchers in the creation of more accurate filters for dialogue data, which could lead to improvements in the quality and efficiency of collecting multi-modal dialogues. We were limited with resources and yet tested model accuracy with a lot of repeats on a subset, that could have led to distribution bias across datasets. As well, low recall means we do not include a lot of valid samples, which reduces our total dataset size. Therefore, the approach presented in this work could facilitate the development of more effective multi-modal models. There is also potential to further improve the size of the dataset through labeling or model upgrades, as mentioned above.

\smallskip

Moreover, it is essential to conduct extensive research to compare multi-modal approaches for solving this task. This involves comparing models using cross-attention and concatenation of embeddings to that task, as well as conducting experiments with different Language Models and Visual Encoders. Such an investigation can lead to the development of more effective and accurate multi-modal models.

\section{Conclusion}

In conclusion, our work presents a new task of interpreting images in the context of dialogue and proposes a novel approach to construct a multi-modal dialogue dataset to tackle this challenge. We utilized a two-stage process that involves identifying utterances that can be replaced with images and selecting relevant images using visual question answering models. Through this process, we created the IMage Augmented multi-modal Dialogue dataset (IMAD), which is validated and labeled, providing a valuable resource for further research in this area. Additionally, we proposed a baseline model trained on IMAD, which outperformed existing models that do not incorporate images. Our work demonstrates the potential of incorporating visual information in dialogue systems and highlights the need for more research in this area. Future work can explore the use of more advanced techniques for identifying relevant images and developing more sophisticated models that can effectively incorporate visual information into dialogue systems.

\bibliographystyle{unsrt}  
\bibliography{references}  

\section*{Appendix A. Implementation details}
Formally pairs of utterance and images are made this way.

Giver set of images $I$, we created pairs $$\forall u \quad \exists \, (u,img): img = \operatorname{argmax}_{i \in I} \{\operatorname{cosine}(\operatorname{emb}_{CLIP}(u), \operatorname{emb}_{CLIP}(i))\}$$, where $\operatorname{emb}_{CLIP}(u)$ is taking embedding from CLIP for utterance and $\operatorname{emb}_{CLIP}(i)$ is taking embedding from CLIP for images and $\operatorname{cosine}(a,b) = \frac{a \cdot b}{ \| a \| \cdot \| b\|}$ is calculating cosine similarity between vectors in a usual sense.

Main features are described in depth below:

\textbf{Image Score}. For utterance $\operatorname{IS}(u)$ is the maximum cosine similarity between utterance and all images embeddings extracted from CLIP. 

$\displaystyle\max_{i \in I} \{\operatorname{cosine}(\operatorname{emb}_{CLIP}(u), \operatorname{emb}_{CLIP}(i))\}$

\smallskip

\textbf{Maximum Entity Score}. We follow the idea, that most of entities in the NER datasets are nouns \cite{Noun-based}. First, each utterance undergoes a noun extraction process and has corresponding noun set $ENT_u = \{noun \, | \, \forall noun \in u \}$.  Second, for each entity in the set Image Score is calculated $\operatorname{IS}(entity)$. That forms set of Image Scores of utterance nouns which we call Entity Scores for utterance $ES_u = \{\operatorname{IS}(entity) \, | \, \forall entity \in ENT_u \}$. Finally, we take maximum of Entity Scores $\operatorname{MES}(u) = \max ES_u$.

\smallskip

\textbf{Sentence Similarity}. It is obtained from comparing image caption and initial utterance. First, for each corresponding image captions are generated $\operatorname{caption}(img)$. We do this with VIT-GPT2 model \cite{kumar2022imagecaptioning}. Then we find similarity between utterance and generated caption with cosine distance between their embeddings from SentenceBert $\operatorname{SS}(u,img) = \operatorname{cosine}(\operatorname{emb}_{SB}(u), \operatorname{emb}_{SB}(\operatorname{caption}(img))$ \cite{reimers-2019-sentence-bert} . 

\smallskip

\textbf{BLEU Score}. BLEU \cite{bleu} metric for only unigrams between generated caption and utterance $\operatorname{BLEU_1}(u, \operatorname{caption}(img))$.

\smallskip

\textbf{Threshold}. Binary feature that shows if utterances features listed above are greater than founded thresholds. We found thresholds for Image Score $t_{IS}$, Sentence Similarity $t_{SS}$ and Maximum Entity Score $t_{MES}$ by maximizing precision on labeled subset $U$ via grid-search on triplets $(t_{IS}, t_{SS}, t_{MES})$. To reduce the computations, each threshold was chosen as k-th statistic in set of train values sorted in descending order. Therefore we were grid searching through k-th statistic for each parameter with step equals 10. Formally, threshold is $THR_{u, img} =  \mathbbm{1} \{\operatorname{SS}(u,img) >= t_{SS}\} \mathbbm{1} \{\operatorname{MES}(u) >= t_{MES}\} \mathbbm{1} \{\operatorname{IS}(u) >= t_{IS}\}$

\smallskip

The hyperparameters were determined through a grid-search that maximized the precision score. The resulting model consisted of 500 estimators with class weights of 5 to 1 for not-replaceable and replaceable, respectively. The model used the Gini criterion, a maximum depth of 2, and the square root of the number of features in each estimator.

\smallskip

Thresholds features were tested on labeled dataset and brute forced on the 10 step grid. We report on listed below thresholds, that results in precision = 0.921171 and recall = 0.068.
    \begin{itemize}
        \item Image Score = 0.33265801843083337
        \item Sentence Similarity = 0.12116438820166667
        \item Maximum Entity Score = 0.3103302687291667
    \end{itemize}

\medskip

There is a list of pictorialization features that were tested, but resulted in worse metrics. 
\begin{itemize}
    \item Embedding representations from SentenceBERT and subsets of embeddings representation
    \item Image-text matching loss from BLIP
    \item Answers from VQA model to the question "does statement *utterance* describe picture well?" transitioned to binary feauture for model output "yes" and "no" 
    \item Answers from VQA model to the question "Can the utterance *utterance* be described by the picture?" transitioned to binary feauture for model output "yes" and "no" 
    \item Cardinality of intersection between nouns from utterance and objects that VQA answers to the question "what objects are in the picture?"
    \item Confidence of model on "The picture shows *utterance*". It was calculated as sum of logarithmic probabilities of tokenized phrase without taking in account pattern phrase "The picture shows".
    \item Text feauteres: Number of parts of speech in utterance, punctuation in utterance, utterance length, number of words, SMOG Index, LIX Index, Flesch–Kincaid readability tests, Coleman–Liau index. Lexical diversity metrics: TTR, RTTR, CTTR, HTTR, STTR, MTTR, DTTR, MATTR, MSTTR, MTLD, MAMTLD, HD-D \cite{ruTS}
\end{itemize}

\begin{table}[pt]
\caption{Metrics for best (in terms of Precision) Random Forest model for predicting if utterance is replaceable}
\label{TableMetricsRF} 
\begin{center}
\begin{tabular}{lccc}
\hline \bf Metric & Mean & Standart Deviation & Median  \\ \hline
Precision &  0.975000 &  0.156125 &  1.000000 \\
Recall &  0.030990 &  0.006374 &  0.031250 \\
F1 &  0.060042 &  0.012072 &  0.060606 \\
\hline
\end{tabular}
\end{center}
\end{table}

\begin{table}[pt]
\caption{Comparison for different ML Algorithms in terms of Precision and Recall}
\label{TableAlgos} 
\begin{center}
\begin{tabular}{lcc}
\hline \bf Algorithm &  Precision &    Recall \\\hline
Random Forest  &   0.975000 &  0.030990 \\
Kernel SVM &   0.958333 &  0.037500 \\
Gradient Boosting  &   0.816667 &  0.026302 \\
KNN &   0.500188 &  0.035677 \\
\hline
\end{tabular}
\end{center}
\end{table}

\section*{Appendix B. Dataset Reasoning}
\textbf{DailyDialog} \cite{Li-Su-2017}. Daily Dialog is a popular source of human-written dialogues, contains 76k utterances with context. Moreover, dialogues stick to the certain topic or object and they end after reasonable turn. These features are essential for building context related models.

\smallskip

\textbf{Mutual dataset} \cite{cui-etal-2020-mutual}. This dataset was crawled from English students books, writtend with expert linguists and contains dialogues with reasonings. Contains 18k utterance with context.

\smallskip

\textbf{Common Sense Dialogues} \cite{Zhou-2021}. This dataset focuses on real-life common sense dialogues. It contains 43k utterances with context from human-written dialogues, collected from assessors on Amazon Mechanical Turk (MTurk).

\smallskip

\textbf{Empathetic Dialogues} \cite{Rashkin-2018}. This dataset was also collected via MTurk. The main goal is to focus on emotional and personal dialogues. It contains 76k utterances with context.

\smallskip

\textbf{Dream Dataset} \cite{Sun-2019}. This dataset contains 14k utterances with context from dialogues from English students books. The goal is to make wide range of dialogues, that would include both common knowledge and reasoning.

\smallskip

\textbf{Persona Chat} \cite{Zhang-2018}. We have utilized Persona Chat because it was collected using assessors and contains 223k utterances with context from small real life dialogues. It was well validated and contains dialogues with personas.

%\section*{Appendix C. Finetuned models details}
%\import{sections/}{appendix_C.tex}

\section*{Appendix C. Training Details}

To evaluate the contribution of visual information into solving the task we finetune two models in different settings: using both image and context as input and replacing image feature vector with constant input. We expect the model which has access to visual modality to perform better, which would mean that images in the composed dataset are actually relevant and useful for utterance prediction.

To train our models we utilise a prefix LM learning paradigm as opposed to usual next token prediction used to finetune BLIP on captioning tasks. We use the image and the last utterance before the image as inputs for the model and aim to predict the substituted utterance. We let the decoder use a square attention mask for the context tokens and a triangular attention mask for the utterance tokens we want to predict.

For both models we freeze the image encoder part during training for efficiency and to reduce compute cost.

\begin{figure}[h]
    \centering
    % TODO: uncomment
    \includegraphics[width=0.49\linewidth]{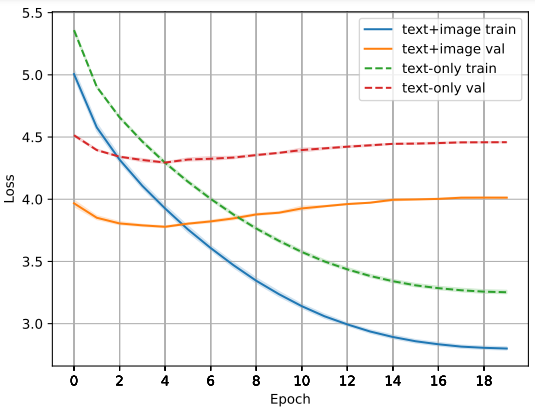}
    \includegraphics[width=0.49\linewidth]{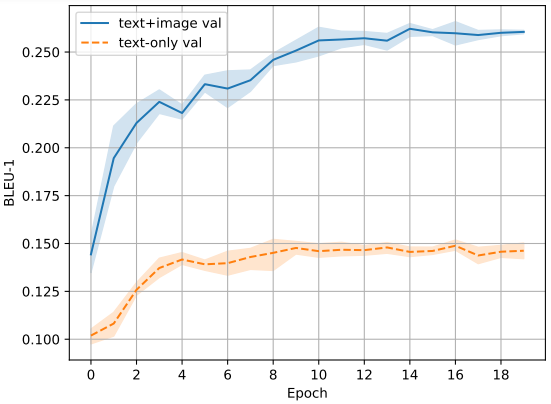}
    \caption{Training metrics: left figure shows train and validation loss for both finetuned BLIP models during training, right figure shows BLEU-1 metrics on validation during training.}
    \label{fig:bleu_val}
\end{figure}

Both models are trained on 4336 samples from the composed dataset, leaving 400 samples for validation, for 20 epochs with batch size of 10 samples, learning rate of 1e-5, and cosine learning rate scheduler with linear warmup (Figure \ref{fig:bleu_val}). For each setting we train 5 models initializing from different seeds for better quality estimation.

\section*{Appendix D. Labeling methodology}
The detailed description for classes are the following:

\begin{enumerate}
    \item \textbf{Perfect Match}. This label is assigned when utterance has only 1 sense, and it is fully transferred with an image. If image could not transfer fully the sense of utterance and it could be done only with context knowledge, then also this label is assigned. There should be no factual mistakes, image should not be specific to cultural differences. The heuristic rule was to question yourself "Could i possibly came to this phrase, knowing image and context?". 
    \item \textbf{Partial Match}. This label is assigned when utterance has 2 or more distinct senses and image transfers one of them fully. It also should not be specific to culture or contain mistakes. In fact rules are the same as for Perfect Match, but applied to one of the senses in the utterances contains 2 or more senses.
    \item \textbf{Undefined}. This label is assigned when image is specific to cultural differences or when image can not transfer one of the senses of the utterance and the context could not help to recover untransferred sense.
    \item \textbf{No Match}. This label is assigned when image contains factual errors about utterance or when none of the entities from utterance present in the image.
\end{enumerate}

\smallskip

\begin{figure}[h]
\centering
% TODO: uncomment
\includegraphics[scale=0.5]{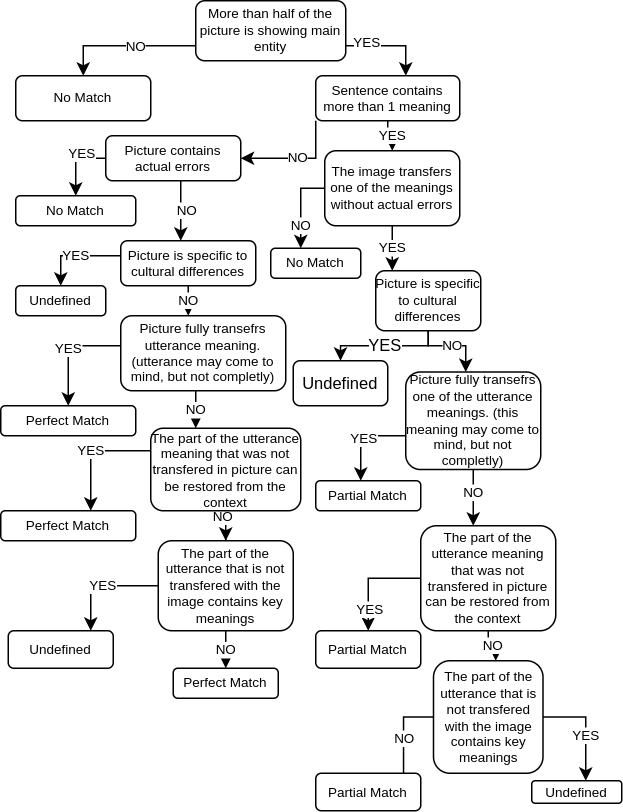}
\caption{Labeling Methodology}
\label{LabelingMethod}
\end{figure}

A decision tree was designed as the primary instruction for the labeling process, which aimed to assist the assessors in assigning appropriate labels to the samples. The decision tree was presented in Figure~\ref{LabelingMethod} and consisted of closed questions at each node, or terminal nodes containing the desired label. The assessors were instructed to follow the decision tree, starting from the root node and answering the questions until the label for each sample was reached. This strategy yielded high inter-rater reliability among the 3 assessors, indicating its efficacy in achieving consistent labeling outcomes.

\end{document}